\documentclass[letterpaper, 10 pt, conference]{ieeeconf}  

\IEEEoverridecommandlockouts                              

\usepackage{amsmath,amssymb,amsfonts}
\usepackage{algorithmic}
\usepackage{graphicx}
\usepackage{textcomp}
\usepackage{wrapfig}
\usepackage{multirow}
\usepackage{booktabs}
\usepackage{cite}
\usepackage{subcaption}
\usepackage{color}
\usepackage[colorlinks=false,pdfborder={0 0 0}]{hyperref}

\newcommand{\fref}[1]
{Fig. \ref{#1}}

\newcommand{\tref}[1]
{Table \ref{#1}}

\title{\LARGE \bf
CLAIM: Camera-LiDAR Alignment with Intensity and Monodepth
}

\author{Zhuo Zhang$^{1}$, Yonghui Liu$^{1}$, Meijie Zhang$^{1,*}$, Feiyang Tan$^{1}$ and Yikang Ding$^{1}$
\thanks{$^{1}$All the authors are with Mach Drive, Beijing, China.
        {\tt\small \{zhuo.zhang04, yonghui.liu, meijie.zhang, feiyang.tan, yikang.ding\}@mach-drive.com}}%
\thanks{$^{*}$Project leader}
}

\begin{document}

\maketitle
\thispagestyle{empty}
\pagestyle{empty}

\begin{abstract}

    In this paper, we unleash the potential of the powerful monodepth model in camera-LiDAR calibration and propose CLAIM, a novel method of aligning data from the camera and LiDAR. Given the initial guess and pairs of images and LiDAR point clouds, CLAIM utilizes a coarse-to-fine searching method to find the optimal transformation minimizing a patched Pearson correlation-based structure loss and a mutual information-based texture loss. These two losses serve as good metrics for camera-LiDAR alignment results and require no complicated steps of data processing, feature extraction, or feature matching like most methods, rendering our method simple and adaptive to most scenes. We validate CLAIM on public KITTI, Waymo, and MIAS-LCEC datasets, and the experimental results demonstrate its superior performance compared with the state-of-the-art methods. The code is available at \url{https://github.com/Tompson11/claim}.

\end{abstract}

\section{INTRODUCTION}
Nowadays, cameras and LiDARs are the most common sensors in both autonomous driving and embedded artificial intelligence systems. To build a smart and robust system, fusing the data of these two complimentary sensors is usually the core task. A good alignment of the LiDAR point cloud and the camera image is essential for valid data fusion since it ensures the consistency of perceived information and makes the fused results more reliable. 

Extrinsic calibration aims at estimating the rigid transformation between the LiDAR and camera coordinate and is the prerequisite for good alignment. Hand-eye calibration \cite{Hand-Eye} is a classic calibration method based on relative motions of the sensors. To obtain accurate results, it requires careful time synchronization and specific motion patterns, which compromises its usability in most scenarios. As another mainstream methodology, cross-modal matching-based methods rely on building correspondences between features in 3D LiDAR point clouds and 2D images and minimizing the final projection distances. The features used for matching vary widely, including patterns on specifically designed targets \cite{Chessboard-points,Chessboard-Planes,ArUco-marker,Wooden-board}, geometric edges \cite{HKU-Mars,Line-zhang}, and semantic instances like lanes \cite{CRLF} and vehicles \cite{Sun,CalibAnything}. These methods usually have complex pipelines involving data processing, feature extraction, and feature matching, and are not robust enough since they are scene-dependent. With the rapid development of deep learning, numerous end-to-end methods regressing extrinsic parameters have emerged and achieved great performance on specific datasets. However, their accuracy can degrade drastically in the face of scenes or sensor configurations different from the training sets.

\begin{figure}[t]
    \centering
    \includegraphics[width=\linewidth]{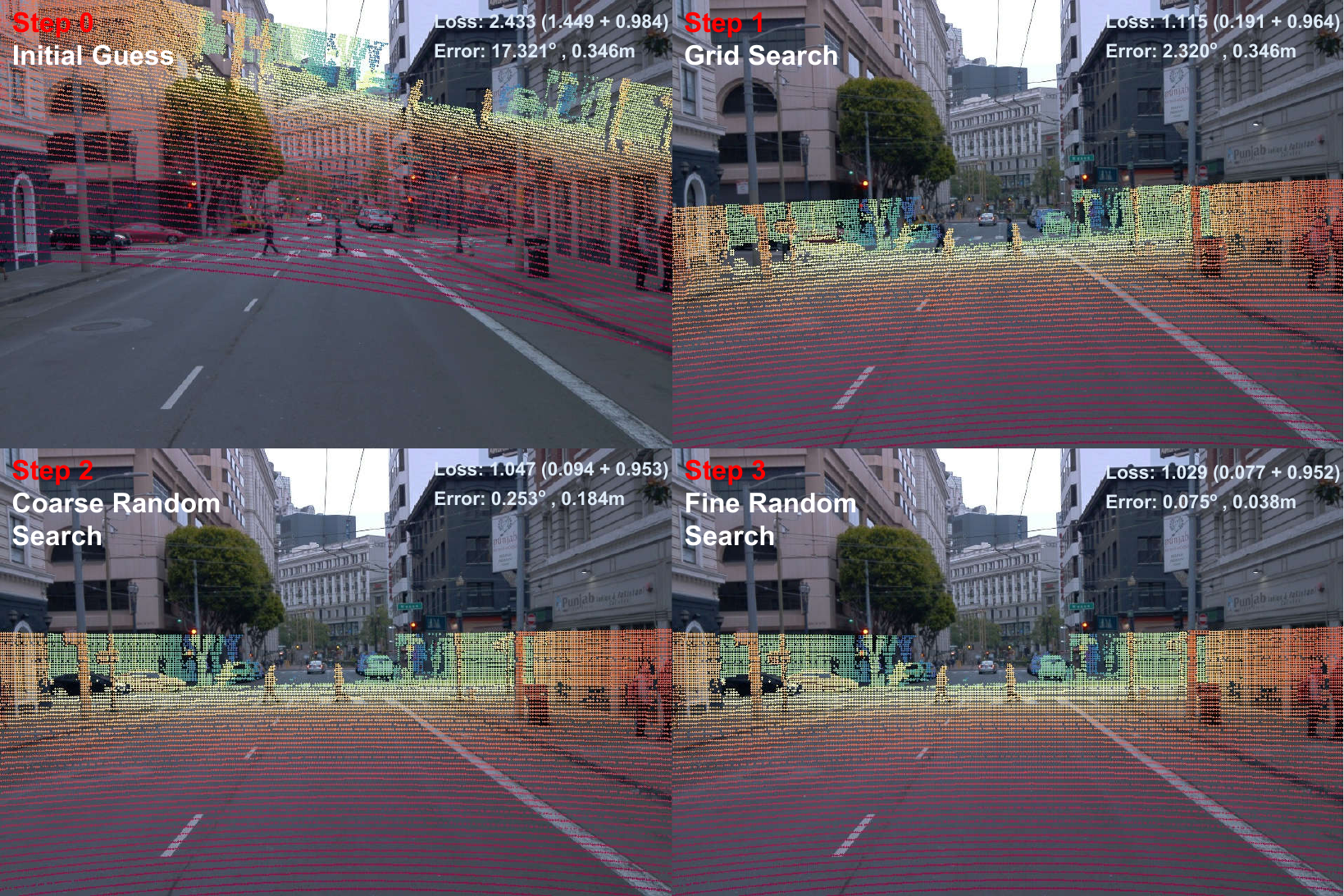}

    \caption{Illustration of our method obtaining a finer camera-LiDAR alignment result step-by-step. The top-left texts mean each step of the pipeline, and the top-right values indicate the values of our defined loss function (structure + texture) and the calibration errors (Euler, translation)  after the step. It is seen that LiDAR points of pedestrians and poles gradually approach the right places as the loss function decreases.}
    \label{fig:waymo_4steps}
\end{figure}

In essence, the main restriction on the camera-LiDAR alignment is the absence of depth information for images. To overcome this problem, some learning-based methods \cite{CalibDepth, CalibNet} incorporate image depth estimation modules into their pipelines and minimize the depth difference between the two sensors. However, they are still weak in generalizability. Recently, monocular depth estimation has become a hot research topic, and large pre-trained monodepth models like \cite{depthv2} show amazing performance. Though monodepth models still struggle to provide accurate absolute depth values (namely metric depth), the relative depth they estimate can already reflect the abundant depth variation and distribution information of the scenes, which exactly can be cross-checked with LiDAR data. So we wonder if we can take good advantage of these powerful models to achieve a good camera-LiDAR alignment.

In this paper, we propose $\textbf{CLAIM}$, a novel $\textbf{C}$amera-$\textbf{L}$iDAR $\textbf{A}$lignment method with $\textbf{I}$ntensiy and $\textbf{M}$onodepth. We employ a patched Pearson correlation-based structure loss w.r.t. the monodepth image and LiDAR depth projection, and a mutual information-based texture loss w.r.t. the grayscale image and LiDAR intensity projection to measure the camera-LiDAR alignment consistency from two aspects. A coarse-to-fine searching method is designed to find the optimal extrinsic parameters minimizing the total loss. The whole pipeline is concise without complicated data processing, feature matching, and extra training steps. To summarize, our main contributions are as follows:
\begin{enumerate}
\item We unleash the potential of powerful monodepth models in camera-LiDAR calibration, proposing a novel calibration method termed CLAIM, which makes full use of information in the grayscale image, monodepth image, LiDAR depth, and LiDAR intensity. 
\item Not dependent on specific targets, curated features, or exhaustive training, the entire pipeline is concise and adaptive to most scenes.
\item Extensive experiments on various public datasets are conducted to validate the accuracy and robustness of our method.
\end{enumerate}
\vspace{+0.1cm}

\begin{figure*}[htbp]
    \centering
    \includegraphics[width=0.9\linewidth]{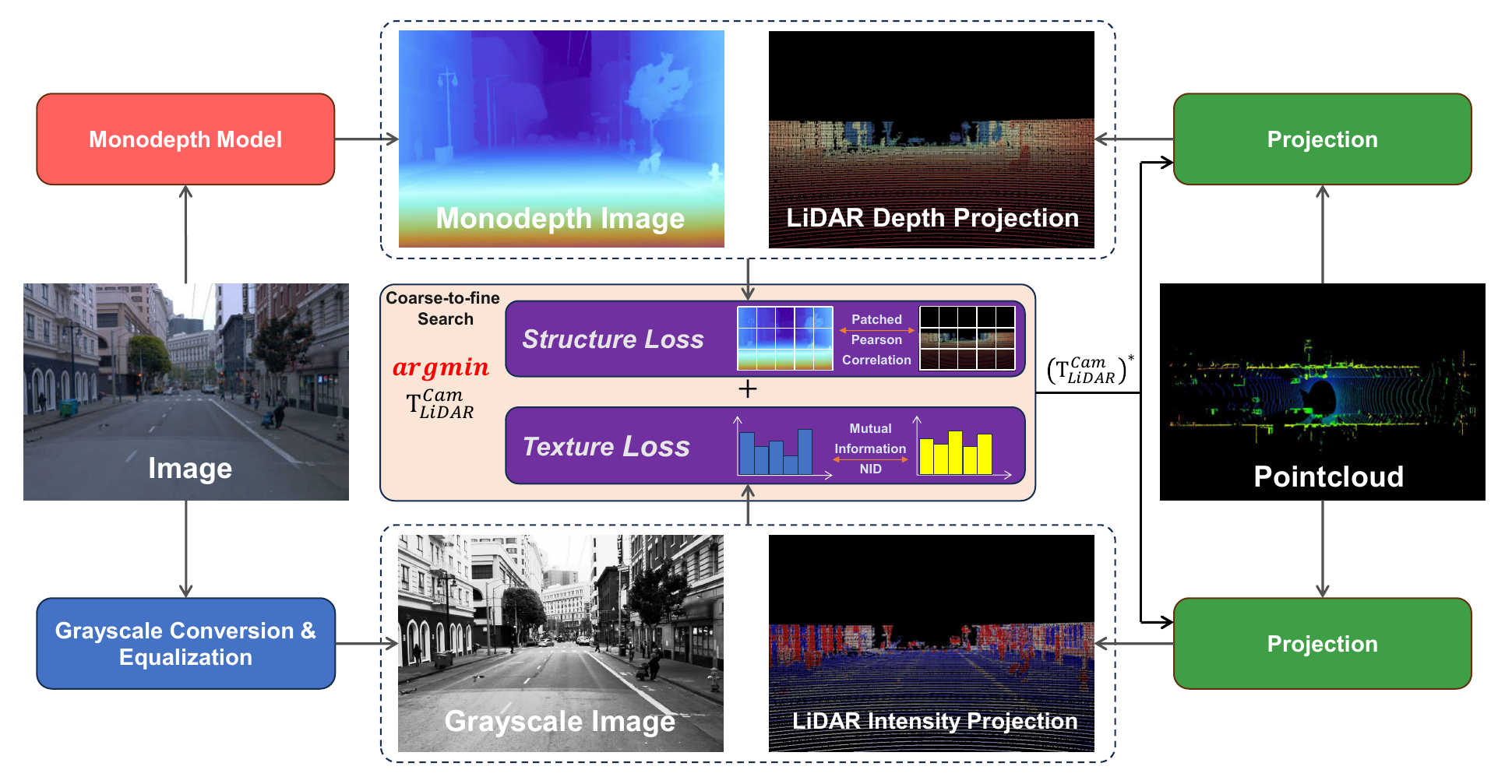}
    \caption{Overview of the proposed CLAIM. First a monodepth image and a grayscale image are generated from the input image by monodepth model and image processing respectively. Then a coarse-to-fine search is conducted to find the extrinsic parameters minimizing the loss function consisting of a structure loss and a texture loss. The structure loss is a patched Pearson correlation-based loss w.r.t the monodepth image and LiDAR depth projection, and the texture loss is a mutual information-based NID loss w.r.t the grayscale image and LiDAR intensity projection.}
    \label{fig:overview}
\end{figure*}

\section{Related Works}
Based on principles, LiDAR-camera calibration methods can be divided into three categories: 1) motion-based; 2) target-based; 3) scene-based.

Motion-based methods work by aligning motion trajectories of the two sensors and are usually termed hand-eye calibration \cite{Hand-Eye}. Works such as \cite{Hand-Eye-method2} have conducted useful explorations on this category of methods. Since motion-based methods do not need to directly fuse the data of two modalities, they can calibrate sensors with little covisibility. However, their accuracy depends on the quality of time-synchronization, individual odometry, and motion patterns, which compromise their usability in most scenarios.

Target-based methods employ specially customized targets with distinctive color, reflectivity, or shape and match the target patterns in images and LiDAR point clouds. The chessboard is the most common target because its corner points \cite{Chessboard-points} are easy to extract in both images and point clouds and are suitable for matching. To increase accuracy, some methods resort to elaborate targets like ArUco makers \cite{ArUco-marker} and wooden boards with holes \cite{Wooden-board}. Due to the predominant features and known sizes of the targets, these methods can usually achieve high accuracy. However, the dependence on targets or human interventions greatly impairs their accessibility and flexibility in practice.

In contrast to target-method methods, scene-based methods rely on the features presented in the scene for alignment. Lines/Edges are rich in real scenes, thus HKU-Mars \cite{HKU-Mars} try to extract and associate lines in two modalities. To enhance feature extraction, some methods turn to semantic segmentation. CRLF \cite{CRLF} uses image segmentation to help extract lines of lanes and poles and match them with lines fitted in the point cloud. \cite{Sun} and \cite{CalibAnything} both focus on objects like vehicles in the scenes and aim at aligning their points to the corresponding semantic masks. MIAS-LCEC \cite{MIAS-LCEC} designs a cross-modal mask matching algorithm to associate the corner points of the segmented RGB image with the segmented LiDAR intensity image. EdO-LCEC \cite{EDO-LCEC} further extends MIAS-LCEC by incorporating depth information and spatial-temporal optimization to improve accuracy. Apart from the aforementioned explicit feature matching-based methods there are also approaches performing a direct alignment of the grayscale image and LiDAR intensity image with mutual information, and the representative works include UMich \cite{UMich} and DVL \cite{DVL}. Compared with the previous two categories of methods, scene-based methods have broader practicality and flexibility. While they usually have complicated pipelines and their performance may vary on different scenes.



\section{Methodology}

\subsection{Overview}
\fref{fig:overview} presents the pipeline of our proposed CLAIM. The system input is a pair of image and point cloud from the target camera and LiDAR respectively. For the image, we use a pre-trained monodepth model (Depth Anything V2 \cite{depthv2}) to get the estimated monodepth image (MI) and meanwhile perform grayscale conversion and equalization to get the grayscale image (GI). For the point cloud, we just perform the equalization on the point intensity as mentioned in \cite{DVL}. Then, a coarse-to-fine search is conducted to estimate the optimal LiDAR-camera extrinsic transformation which minimizes the loss function consisting of a structure loss and a texture loss. The structure loss is a patched Pearson correlation-based loss calculated with MI and LiDAR depth projection (LDP), which measures the structure similarity. The texture loss is a mutual information-based NID loss calculated with GI and LiDAR intensity projection (LIP), which reflects the texture consistency. These two losses evaluate the matching results of two modalities in different aspects and can serve as an appropriate alignment guideline.

Now we define notations used throughout the paper. Given the input color image $\mathbf{I}$ with height $H$ and width $W$, we use $\mathbf{I}_{mono},  \mathbf{I}_{gray} \in \mathbb{R}^{H \times W} $ to denote the corresponding MI and GI. Extrinsic between the LiDAR and the camera is $\mathbf{T}_L^C = [\mathbf{R}_L^C, \mathbf{t}_L^C]$, where $\mathbf{R}_L^C \in \mathbb{R}^{3\times3}, \mathbf{t}_L^C \in \mathbb{R}^{3}$
are the rotation matrix and translation vector. Assume camera intrinsic matrix is known and represented as $\mathbf{K} \in \mathbb{R}^{3\times3}$, for each 3D point $\mathbf{p}_i^{L} \in \mathbb{R}^{3}$ in the point cloud we can use \eqref{equ:projection} to project it onto image plane and set the value of the corresponding pixel $(u_i, v_i)$ to point depth $(\mathbf{p}_i^{C})_z$ for LDP $\mathbf{L}_{dep} \in \mathbb{R}^{H \times W}$ or point intensity for LIP $ \mathbf{L}_{int} \in \mathbb{R}^{H \times W}$.
\begin{equation}
\begin{aligned}
    \mathbf{p}_i^{C} = \mathbf{R}_L^C\mathbf{p}_i^L + \mathbf{t}_L^C\ \\
    [u_i, v_i, 1]^T = \mathbf{K} \mathbf{p}_i^{C} / (\mathbf{p}_i^{C})_z
    \label{equ:projection}
\end{aligned}
\end{equation}

\subsection{Loss Function} 
LiDAR-camera calibration or alignment task always boils down to defining an appropriate metric for modality consistency and finding the optimal transformation. To make the whole method simple, general, and robust, we introduce a loss function consisting of a structure loss and a texture loss.

Our structure loss is inspired by SparseGS \cite{SparseGS}, in which the authors use monodepth to regulate the geometry of 3D gaussian-splatting (3DGS \cite{3DGS}) in a novel way. Specifically, It randomly samples some non-overlapping patches of monodepth image and dense depth image rendered by 3DGS to calculate Pearson loss at each iteration, which can help maintain the local spatial structure and resist the variation of scale and shift parameters of monodepth at different locations. In our method, we use a similar idea to measure the structural similarity between MI and LDP, defining the structure loss as follows:
\begin{equation}
    \mathcal{L}_{structure}^{(u_0, v_0, S)} = \frac{1}{N_{valid}}\sum_{i = 1}^{N_H}\sum_{j = 1}^{N_W}1 - SPCC(\mathbf{I}_{mono}^{i,j}, \mathbf{L}_{dep}^{i,j})
    \label{equ:structure_loss}
\end{equation}
\begin{equation}
    \scalebox{0.6}{$SPCC(X, Y)$} = \begin{cases}
          1 & \scalebox{0.6}{$|\mathcal{V}_Y| < P$} \\
          \frac{|\mathcal{V}_Y|\sum\limits_{\scalebox{0.4}{$k\in\mathcal{V}_Y$}}{\!\!X_kY_k} - \bar{X}\bar{Y}}
          {
            \sqrt{{|\mathcal{V}_Y|\sum\limits_{\scalebox{0.4}{$k\in\mathcal{V}_Y$}}{\!\!X_k^2}} - \bar{X}^2}
            \sqrt{{|\mathcal{V}_Y|\sum\limits_{\scalebox{0.4}{$k\in\mathcal{V}_Y$}}{\!\!Y_k^2}} - \bar{Y}^2}
          }
          & \scalebox{0.6}{$|\mathcal{V}_Y| \geq P$}
       \end{cases}
       \label{equ:SPCC}
\end{equation}
\begin{equation}
    \bar{X} = \sum_{k\in\mathcal{V}_Y}{X_k}, \quad \bar{Y} = \sum_{k\in\mathcal{V}_Y}{Y_k}
\end{equation}
where $\mathcal{L}_{structure}^{(u_0, v_0, S)}$ means the structure loss is calculated for tightly-packed patches starting from the pixel at $(u_0, v_0)$ with size $S\!\times\!S$. $N_H = \left \lfloor \frac{H - v0}{S} \right \rfloor$, $N_W = \left \lfloor \frac{W - u0}{S} \right \rfloor$ are the patch numbers in vertical and horizontal direction. $\mathbf{X}^{i,j}$ represents the patch of $\mathbf{X}$ spanning from top-left pixel $\left (u_0 + (j - 1) \cdot
 S, v_0 + (i - 1) \cdot S \right )$ to bottom-right pixel $\left (u_0 + j \cdot S - 1, v_0 + i \cdot S - 1 \right )$. $SPCC(X, Y)$ denotes the sparse Pearson correlation coefficient of $X$ and $Y$, which falls in $[-1, 1]$ and represents the extent of linear correlation. 
 
 It is noteworthy that $SPCC(X, Y)$ assumes $Y$ is sparse and $ \mathcal{V}_Y $ is the set containing all pixels with non-idle values. LDP always corresponds to $Y$ in $SPCC$ since the point cloud may be sparse and cannot generate a dense LDP. We find that $SPCC$ tends to be inaccurate on extremely sparse patches, thus a threshold $P$ is set to distinguish valid patches for the structure loss. Only the LDP patches with more than $P$ non-idle values can contribute to the loss and their total number is denoted as $N_{valid}$. 

 Since the structure loss encourages a consistent depth distribution of MI and LDP, it actually embodies the idea of aligning edges (i.e. locations with great depth variation) and planes (i.e. locations with low depth variation) in the point cloud and image, which however usually requires careful steps of feature extraction and matching in the previous methods. However, the structure loss may be unreliable when the estimated monodepth is not accurate enough, so we introduce the texture loss in \cite{DVL} to improve robustness:
 \begin{equation}
    \mathcal{L}_{texture} = \text{NID}(\mathbf{I}_{gray}, \mathbf{L}_{int}) = 1 - \frac{\text{MI}(\mathbf{I}_{gray}; \mathbf{L}_{int})}{\text{H}(\mathbf{I}_{gray},  \mathbf{L}_{int})}
    \label{equ:NID}
\end{equation}
\vspace{-0.3cm}
\begin{equation}
    \text{MI}(X; Y) = \text{H}(X) + \text{H}(Y) - \text{H}(X, Y)
    \label{equ:mutal_information}
\end{equation}
\vspace{-0.5cm}
\begin{equation}
    \text{H}(X) = \sum_{x \in X_{bin}}p(x) \log p(x)
    \label{equ:entropy}
\end{equation}

 As \eqref{equ:NID}, the texture loss is the normalized information distance (NID) between GI and LIP. $\text{MI}(X; Y)$ means the mutual information of $X$ and $Y$, which equals the sum of marginal entropy $H(X), H(Y)$ minus the joined entropy $H(X, Y)$ and reflects the associataion between $X$ and $Y$. Note that the subscript $bin$ in \eqref{equ:entropy} indicates the entropy is calculated with a histogram and $p(\cdot)$ denotes the bin value. Similar to $SPCC$, $Y$ in the \text{NID}(X, Y) always corresponds to sparse LIP, thus the calculation of \eqref{equ:entropy} only considers the non-idle values of $Y$. The texture loss assumes grayscale values of the image indicate the reflectivity of light, which thus can be correlated with the perceived intensity of the LiDAR. It takes advantage of environmental texture information and is complementary to the structure loss.

Now we can define the calibration task as the optimization problem in \eqref{equ:opt-problem}, where $\lambda_1, \lambda_2$ are weights for the corresponding loss item. Particularly, we use two structure loss items with different and overlapped patches to better capture the spatial structure and improve accuracy.
\begin{equation}
    \arg\min\limits_{\mathbf{T}_L^C} \lambda_1 \left (\mathcal{L}_{structure}^{(0, 0, S)} + \mathcal{L}_{structure}^{(\frac{S}{2}, \frac{S}{2}, S)} \right ) + \lambda_2 \mathcal{L}_{texture}
    \label{equ:opt-problem}
\end{equation}

\subsection{Coarse-to-fine Search}
To solve \eqref{equ:opt-problem}, we adopt a heuristic coarse-to-fine searching method resembling the one in \cite{CalibAnything}.

Given the initial guess $\mathbf{T}_0 = [\mathbf{R}_0(\alpha_0, \beta_0, \gamma_0), \mathbf{t}_0]$ where $\mathbf{R}_0(\alpha_0, \beta_0, \gamma_0)$ is the rotation matrix of $\mathbf{T}_0$ corresponding to Euler angles $\alpha_0, \beta_0, \gamma_0$, we first perform a grid search on rotations if the initial rotational error is considerable. Specifically, we test all the perturbation combinations of three Euler angles within 
$[-A, A]$ with $1^{\circ}$ resolution, taking $\mathbf{R}(\alpha_0+\delta\alpha^*, \beta_0+\delta\beta^*, \gamma_0+\delta\gamma^*)$ with the minimum loss in \eqref{equ:opt-problem} as the modified initial rotation guess $\mathbf{R}_1$, and $\mathbf{t}_1 = \mathbf{t}_0$.

Then, a coarse random search is performed to optimize $\mathbf{T}_1$. Let's denote the best result after $k - 1$ search iterations as $\mathbf{T}_1^{*} = [\mathbf{R}_1^{*}(\alpha^*, \beta^*, \gamma^*), \mathbf{t}_1^*]$. At the $k$-th iteration, we generate 216 candidates $\{\mathbf{T}^{i} = [\mathbf{R}(\alpha^* + \delta\alpha^i, \beta^* + \delta\beta^i, \gamma^* + \delta\gamma^i), \mathbf{t}_1+\delta \mathbf{t}^i] \; \big| \; i = 0, 1, \dots, 255\}$ and update $\mathbf{T}_1^*$ with the best one amongst them if it obtains a smaller loss than $\mathbf{T}_1^{*}$. Here $\delta\alpha^i, \delta\beta^i, \delta\gamma^i$ are selected from $\{-0.5, -0.2, -0.1, 0.1, 0.2, 0.5\}$ and thus have a total of $6^3=216$ combinations. Also, we make sure each perturbation angle 
$\delta\theta$ meets $\delta\theta^i = -\delta\theta^{i-128} $ for $i > 128$. The first 128 $\delta\mathbf{t}^i$ are randomly sampled from $[-B, B]^3$ and the last 128 translation perturbations meet $\delta\mathbf{t}^i = \delta\mathbf{t}^{i-128}$. The design scheme of candidates considers rotations in various directions and ensures the translation is identical for a pair of inverse rotations, which helps determine the right rotation direction quickly since the loss is more sensitive to it. Also note that the translation perturbation is always applied on the initial value $\mathbf{t}_1$ instead of the optima $\mathbf{t}_1^*$ since we find this practice is more robust. 

After certain iterations of coarse random search, we can get the estimation $\mathbf{T}_2=\mathbf{T}_1^{*}$. To further improve accuracy, a fine random search is conducted. Its process is identical to the coarse random search only except that the perturbation angles are selected from a smaller set $\{-0.1, -0.04, -0.02, 0.02, 0.04, 0.1\}$. We take the final search result $\mathbf{T}_3$ as the calibration result $\mathbf{T}_L^C$.

\section{Experiment}

\subsection{Experimental Setup and Implimentation Details}

We validate the proposed CLAIM on the various public datasets including KITTI \cite{KITTI}, Waymo \cite{Waymo} and MIAS-LCEC-TF360 \cite{MIAS-LCEC}. We only use the left RGB camera for KITTI, and the top LiDAR and the front camera for Waymo. Depth-Anything-V2-Large \cite{depthv2} is used to get monodepth images. Since its output is inverse depth, the LDP also fills the inverse depth. We set the patch size $S$ to 40, 60, and 80 respectively for KITTI, MIAS-LCEC-TF360, and Waymo according to their image sizes. The threshold $P$ for valid patches is set to 15, and the loss weights $\lambda_1 = 0.2$, $\lambda_2 = 1.0$. Total iterations for coarse and fine random search are both 150, and we use CUDA to accelerate the loss calculation in each iteration.

To evaluate the calibration accuracy and align the metrics of different papers, we use Euler component error $\mathbf{e}_r$ and its magnitude $e_r$ for rotation, translation component error $\mathbf{e}_t^+, \mathbf{e}_t^-$ and their magnitudes $e_t^+, e_t^-$ for translation. The detailed formulars can refer to \eqref{equ:metric}, where $\mathbf{R}_L^C, \mathbf{t}_L^C$ are the ground truth, $\mathbf{\hat{R}}_L^C, \mathbf{\hat{t}}_L^C$ are the estimations, and $\mathbf{r}_L^C, \hat{\mathbf{r}}_L^C \in \mathbb{R}^{3} $ are the Euler angle vectors w.r.t. rotation matrix $\mathbf{R}_L^C, \mathbf{\hat{R}}_L^C$. 

\begin{equation}
    \begin{aligned}
        \mathbf{e}_r &= \left |\mathbf{r}_L^C - \hat{\mathbf{r}}_L^C \right |\\
        \mathbf{e}_t^+ &= \left | \mathbf{t}_L^C - \hat{\mathbf{t}}_L^C \right |\\
        \mathbf{e}_t^- &= \left | (\mathbf{R}_L^C)^{-1} \mathbf{t}_L^C - (\hat{\mathbf{R}}_L^C)^{-1} \hat{\mathbf{t}}_L^C\right |
        \label{equ:metric}
    \end{aligned}
\end{equation}

\begin{table}[]
    \vspace{+0.35cm}
    \renewcommand{\arraystretch}{1.2}
    
    \caption{Mean errors of different methods on the KITTI and Waymo dataset. The best results are shown in bold.}
    \label{table:Exp1}

    \resizebox{\linewidth}{!}{
    \begin{tabular}{cccccccc}
    \hline
    Dataset & Methods & Roll($^{\circ}$) & Pitch($^{\circ}$) & Yaw($^{\circ}$) & X($m$) & Y($m$) & Z($m$) \\ \hline
    \multirow{7}{*}{KITTI} & Hand-eye & 0.727 & 0.988 & 0.841 & 0.121 & 0.242 & 0.174 \\
     & HKU-Mars & 0.423 & 0.411 & 0.325 & 0.063 & 0.051 & 0.065 \\
     & Zhu & 0.305 & 0.421 & 0.301 & 0.053 & 0.047 & 0.042 \\
     & Sun & \textbf{0.171} & 0.202 & 0.211 & \textbf{0.028} & \textbf{0.026} & \textbf{0.035} \\
     & CLAIM & 0.280 & 0.240 & 0.167 & 0.054 & 0.048 & 0.068 \\
     & CLAIM-4F & 0.247 & 0.193 & 0.112 & 0.045 & 0.042 & 0.060 \\
     & CLAIM-4F* & \textbf{0.171} & \textbf{0.113} & \textbf{0.079} & 0.031 & 0.038 & 0.041 \\ \hline
    \multirow{7}{*}{Waymo} & Hand-eye & 0.814 & 0.632 & 1.123 & 0.121 & 0.189 & 0.202 \\
     & HKU-Mars & 0.397 & 0.487 & 0.472 & 0.089 & 0.092 & 0.073 \\
     & Zhu & 0.476 & 0.422 & 0.387 & 0.093 & 0.075 & 0.087 \\
     & Sun & 0.291 & 0.221 & 0.260 & 0.048 & 0.029 & \textbf{0.041} \\
     & CLAIM & 0.082 & 0.204 & 0.066 & 0.063 & 0.037 & 0.043 \\
     & CLAIM-4F & 0.124 & 0.209 & 0.063 & 0.057 & 0.033 & 0.045 \\
     & CLAIM-4F* & \textbf{0.073} & \textbf{0.165} &\textbf{0.040} & \textbf{0.043} & \textbf{0.026} & 0.042 \\ \hline
    \end{tabular}
    }
    \vspace{-0.25cm}
\end{table}

\begin{table*}[]
    \vspace{+0.3cm}
    \renewcommand{\arraystretch}{1.2}
    
    \caption{Mean calibration errors of different methods on the KITTI odometry dataset (00-09 sequences). SF and MF indicates methods are single-frame or multi-frame. The best results are shown in bold.}
    \label{table:Exp2}

    \resizebox{\linewidth}{!}{%
    \begin{tabular}{cc|cc|cc|cc|cc|cc|cc|cc|cc|cc|cc}
        \hline
        \multicolumn{2}{c|}{\multirow{2}{*}{Methods}} & \multicolumn{2}{c|}{00}         & \multicolumn{2}{c|}{01}         & \multicolumn{2}{c|}{02}         & \multicolumn{2}{c|}{03}         & \multicolumn{2}{c|}{04}         & \multicolumn{2}{c|}{05}         & \multicolumn{2}{c|}{06}         & \multicolumn{2}{c|}{07}         & \multicolumn{2}{c|}{08}         & \multicolumn{2}{c}{09}          \\ \cline{3-22} 
        \multicolumn{2}{c|}{}                         & $e_{r}(^{\circ})$        & $e_{t}^-(m)$      & $e_{r}(^{\circ})$        & $e_{t}^-(m)$      & $e_{r}(^{\circ})$        & $e_{t}^-(m)$      & $e_{r}(^{\circ})$        & $e_{t}^-(m)$      & $e_{r}(^{\circ})$        & $e_{t}^-(m)$      & $e_{r}(^{\circ})$        & $e_{t}^-(m)$      & $e_{r}(^{\circ})$        & $e_{t}^-(m)$      & $e_{r}(^{\circ})$       & $e_{t}^-(m)$      & $e_{r}(^{\circ})$        & $e_{t}^-(m)$      & $e_{r}(^{\circ})$        & $e_{t}^-(m)$      \\ \hline
        \multirow{6}{*}{SF}        & CRLF             & 0.629          & 4.118          & 0.623          & 7.363          & 0.632          & 3.642          & 0.845          & 6.007          & 0.601          & 0.372          & 0.616          & 5.961          & 0.615          & 25.762         & 0.606          & 1.807          & 0.625          & 5.376          & 0.626          & 5.133          \\ 
                                   & UMich            & 4.161          & 0.321          & 2.196          & 0.305          & 3.733          & 0.331          & 3.201          & 0.316          & 2.086          & 0.348          & 3.526          & 0.356          & 2.914          & 0.353          & 3.928          & 0.368          & 3.722          & 0.367          & 3.117          & 0.363          \\ 
                                   & HKU-Mars         & 33.84          & 6.355          & 20.73          & 3.770          & 32.95          & 12.70          & 21.99          & 3.493          & 4.943          & 0.965          & 34.42          & 6.505          & 25.20          & 7.437          & 33.10          & 7.339          & 26.62          & 8.767          & 20.38          & 3.459          \\
                                   & DVL              & 122.1          & 5.129          & 112.0          & 2.514          & 120.6          & 4.285          & 124.7          & 4.711          & 113.5          & 4.871          & 123.9          & 4.286          & 128.9          & 5.408          & 124.7          & 5.279          & 126.2          & 4.461          & 116.7          & 3.931          \\ 
                                   & MIAS-LCEC        & 5.385          & 1.014          & 0.621          & 0.300          & 0.801          & 0.327          & 1.140          & 0.324          & 0.816          & 0.369          & 4.768          & 0.775          & 2.685          & 0.534          & 11.80          & 1.344          & 5.220          & 0.806          & 0.998          & 0.432          \\ 
                                   & CLAIM            & 0.366          & 0.080          & 0.500          & 0.198          & 0.362          & 0.096          & 0.410          & 0.111          & 0.321          & 0.094          & 0.461          & 0.097          & 0.449          & 0.100          & 0.420          & 0.093          & 0.494          & 0.101          & 0.430          & 0.100          \\ \hline
        \multirow{3}{*}{MF}        & EdO-LCEC         & 0.295          & 0.082          & 2.269          & 0.459          & 0.561          & 0.142          & 0.737          & 0.137          & 1.104          & 0.339          & 0.280          & 0.093          & 0.485          & 0.124          & \textbf{0.188} & 0.076          & 0.352          & 0.115          & 0.386          & 0.120          \\ 
                                   & CLAIM-4F         & 0.206          & 0.047          & 0.268          & 0.115          & 0.207          & 0.056          & 0.235          & 0.065          & 0.234          & 0.055          & 0.315          & 0.060          & \textbf{0.282} & \textbf{0.053} & 0.283          & 0.056          & 0.336          & 0.063          & 0.283          & 0.054          \\ 
                                   & CLAIM-4F*        & \textbf{0.170} & \textbf{0.039} & \textbf{0.195} & \textbf{0.073} & \textbf{0.164} & \textbf{0.047} & \textbf{0.203} & \textbf{0.054} & \textbf{0.190} & \textbf{0.044} & \textbf{0.258} & \textbf{0.046} & 0.312          & 0.062          & 0.242          & \textbf{0.048} & \textbf{0.298} & \textbf{0.054} & \textbf{0.232} & \textbf{0.046} \\ \hline
        \end{tabular}
    }
    \end{table*}

    \begin{figure*}[htbp]
        \begin{subfigure}[t]{0.33\linewidth}
            \centering
            \includegraphics[width=1.0\linewidth]{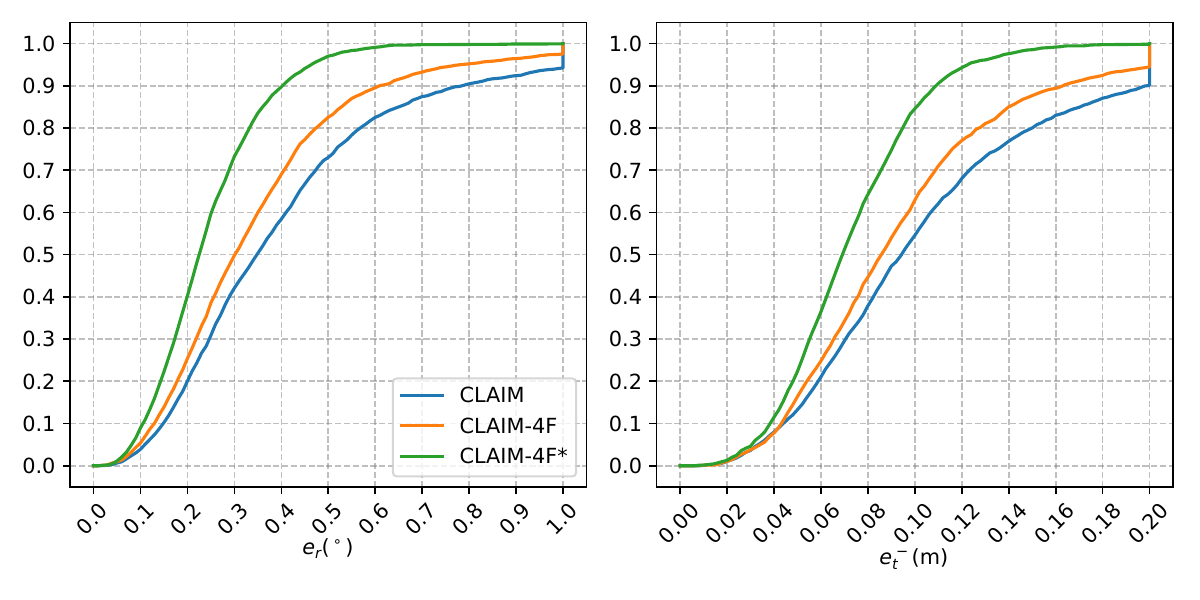}
            \vspace{-0.6cm}
            \caption{KITTI Raw}
            \label{fig:cdf_kitti_raw}
        \end{subfigure}
        \hspace{-0.2cm}
        \begin{subfigure}[t]{0.33\linewidth}
            \centering
            \includegraphics[width=1.0\linewidth]{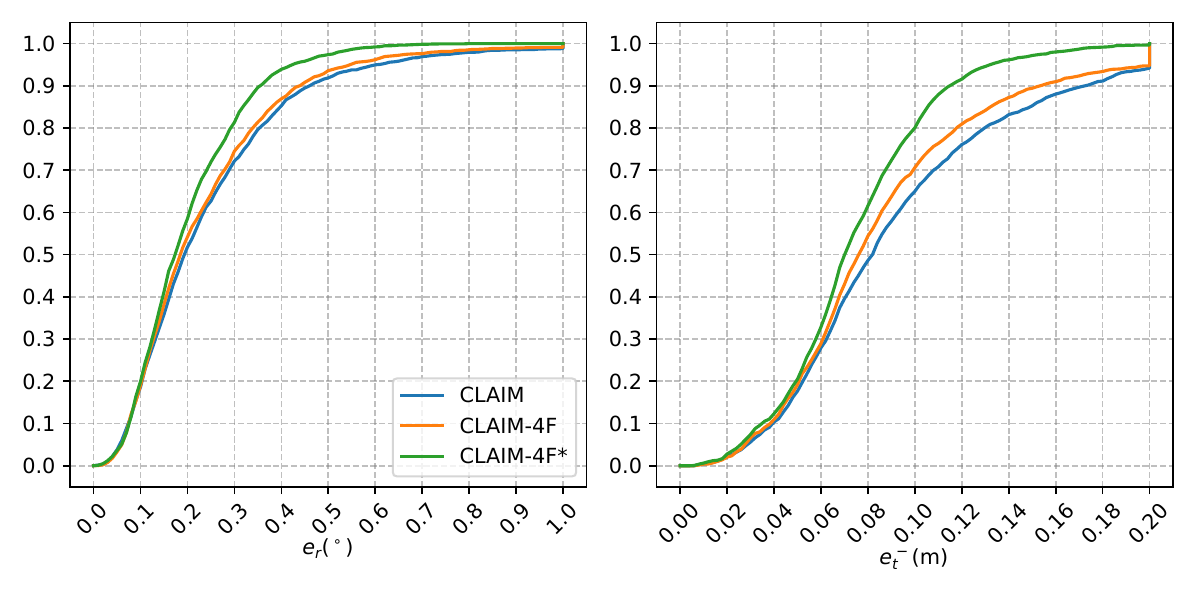}
            \vspace{-0.6cm}
            \caption{Waymo}
            \label{fig:cdf_waymo}
        \end{subfigure}
        \hspace{-0.2cm}
        \begin{subfigure}[t]{0.33\linewidth}
            \centering
            \includegraphics[width=1.0\linewidth]{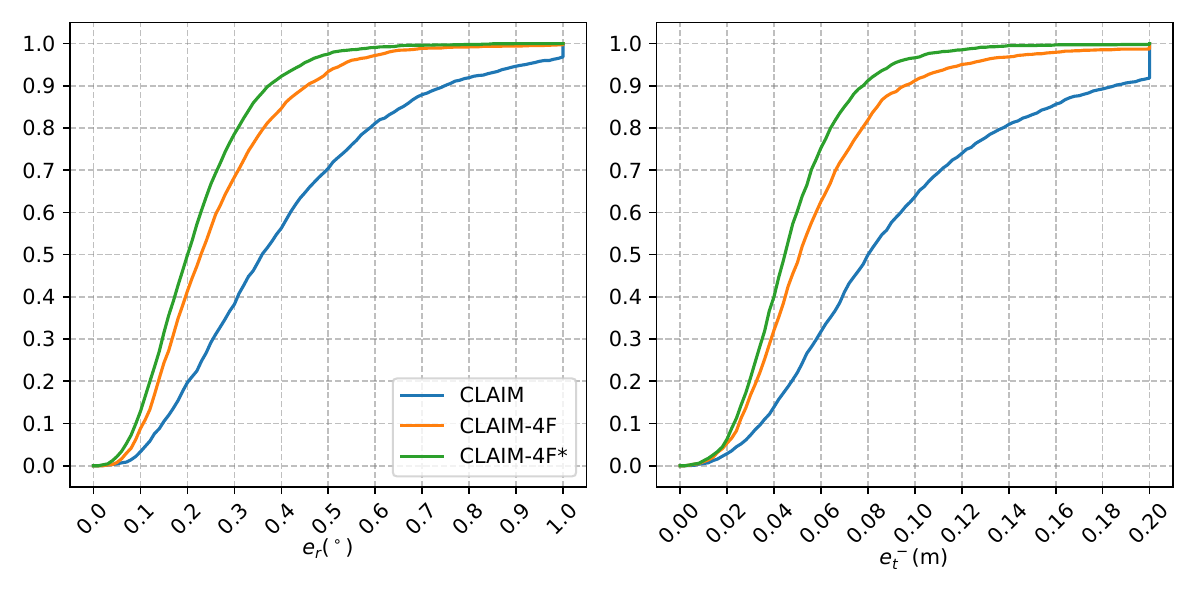}
            \vspace{-0.6cm}
            \caption{KITTI Odometry}
            \label{fig:cdf_kitti_odometry}
        \end{subfigure}
        \caption{Cumulative distribution functions (CDF) of euler errors $e_r$ and translation errors $e_t^-$ for our method on the KITTI raw, Waymo and KITTI odometry datasets. Note that we truncate the $e_r$, $e_t^-$ to $1.0^{\circ}$ and $0.2\mathrm{m}$ respectively, thus the corresponding CDFs may exist leaps at the truncated values.}
        \label{fig:cdf}
        \vspace{-0.3cm}
    \end{figure*}

\subsection{Experimental Results Given a Rough Initial Guess}
\label{sec:exp_rough_initial}
We first evaluate the performance of CLAIM given a rough initial guess. We add $10^{\circ}$ and $0.2m$ respectively to each component of ground truth Euler angles and translation vectors to generate the rough initial guess. The grid search range $A$ for rotation and random search range $B$ for translation are set to $15^{\circ}$ and $0.2m$. The experimental data include 10 sequences from KITTI raw data (2011\_09\_26\_drive\_0005, 0009, 0011, 0018, 0020, 0022, 0035, 0039, 0046 and 0095) and 10 scenarios from Waymo (the first 10 scenarios under v1.3.2 individual-files/testing directory). 

We compare our method with the state-of-the-art Sun's semantic instance-based method \cite{Sun}, Zhu's semantic-based method \cite{Zhu}, and edge-based method HKU-Mars \cite{HKU-Mars}. Since these methods can utilize multiple frames, we also implement two versions of CLAIM with 4 frames namely CLAIM-4F and CLAIM-4F*. CLAIM-4F means we apply a sliding window with length 4 and step 1 on the time-ordered frame list of each sequence to retrieve 4 time-continuous frames for a single calibration, and the loss function in \eqref{equ:opt-problem} incorporates the 4 frames. While CLAIM-4F* uses a randomly shuffled frame list, meaning the retrieved 4 frames are probably time-discontinuous and may correspond to diverse scenes.

\tref{table:Exp1} lists the overall results, where the results of other methods are from \cite{Sun} and the Hand-eye means the hand-eye calibration results used by \cite{Sun} to obtain the initial guess. The translation errors used here are $\mathbf{e}_t^+$. It can be seen that our multi-frame CLAIM-4F and CLAIM-4F* show superior performance of rotation calibration to other methods on both datasets, and the single-frame CLAIM also presents competitive results. Particularly, CLAIM-4F* achieves the minimum errors for all Euler angles (all less than $0.2^{\circ}$), which confirms that incorporating data from diverse scenes is beneficial for calibration since they can provide abundant features. We also notice that the yaw errors of CLAIM-4F* are less than $0.1^{\circ}$, only 37\% and 16\% of Sun's method (i.e. the second best method except ours). This can be attributed to our structure loss taking good advantage of the horizontal depth variation information common in the scenes. For the translation calibration, CLAIM-4F* performs best on Waymo and is slightly behind Sun's method on KITTI, which we think is acceptable in view of our initial guess being rougher compared with the hand-eye results.

\subsection{Experimental Results Given a Fine Initial Guess}
\label{sec:exp_fine_initial}
We also care about the performance of CLAIM when the initial guess is not too bad. In most cases like KITTI and MIAS-LCEC-TF360, the LiDAR coordinate is forward-left-up (FLU) and the camera coordinate is right-down-forward (RDF), thus methods such as DVL \cite{DVL}, MIAS-LCEC \cite{MIAS-LCEC} and EdO-LCEC \cite{EDO-LCEC} set the initial yaw, pitch and roll to $90^{\circ}, 0^{\circ}, 90^{\circ}$ respectively and initial translation to $\mathbf{0}$. For comparison with them, we adopt the same initial guess and only conduct random searches for CLAIM. The experimental data here are 10 sequences from KITTI odometry (00-09) and 2 groups from MIAS-LCEC-TF360 (indoor and outdoor).

\tref{table:Exp2} summarizes the results on KITTI, where the values for other methods are directly from \cite{EDO-LCEC}. It is clearly seen that our CLAIM significantly outperforms other single-frame methods on all sequences, and even shows obvious superiority to the multi-frame method EdO-LCEC on 01-04. Benefiting from fusing multiple frames, CLAIM-4F and CLAIM-4F* further improve the performance of CLAIM and they both outperform EdO-LCEC on the whole. Particularly, CLAIM-4F* attains the smallest translation errors on all sequences and the smallest rotation errors on 9 of 10 sequences among all methods. Moreover, $e_r$ and $e_t^-$ of CLAIM-4F* are all less than $0.32^{\circ}$ and $0.08m$, confirming its accuracy and robustness in various scenes. 

The results on MIAS-LCEC-TF360 are listed in \tref{table:Exp3}. Since each sequence in this dataset contains only one frame and the extrinsics differ between sequences, we implement CLAIM-4F by dividing the provided dense accumulated point cloud into 4 parts to form 4 fake frames for each sequence. CLAIM-4F* is omitted for the lack of diverse scenes within a sequence. Again, CLAIM outperforms all single-frame methods and CLAIM-4F performs best.

Moreover, \fref{fig:cdf} displays the cumulative distribution functions of calibration errors for our method on the three datasets. It indicates that CLAIM has a good adaptability to various scenes and can achieve the accuracy of $0.4^\circ$ and $0.10m$ for around 60\% frames of these datasets. CLAIM-4F* further increase this ratio to around 80\% and significantly decreases the ratio of large errors.    
\begin{table}[]
    \vspace{+0.35cm}
    \centering
    \renewcommand{\arraystretch}{1.2}
    \caption{Mean calibration errors of different methods on the MIAS-LCEC-TF360 dataset. SF and MF indicate the methods are single-frame or multi-frame. The best results are shown in bold.}
    \label{table:Exp3}

    \begin{tabular}{cc|cc|cc}
    \hline
    \multicolumn{2}{c|}{\multirow{2}{*}{Methods}} & \multicolumn{2}{c|}{Indoor}                        & \multicolumn{2}{c}{Outdoor}                           \\
    \multicolumn{2}{c|}{}                         & $e_{r}\left(^{\circ}\right)$ & $e_{t}^-(\mathrm{m})$ & $e_{r}\left({ }^{\circ}\right)$ & $e_{t}^-(\mathrm{m})$ \\ \hline
    \multirow{6}{*}{SF}        & CRLF             & 1.479                        & 13.241              & 1.442                           & 0.139               \\
                               & UMich            & 1.510                        & 0.221               & 6.522                           & 0.269               \\
                               & HKU-Mars         & 85.834                       & 7.342               & 35.383                          & 8.542               \\
                               & DVL              & 39.474                       & 0.933               & 65.571                          & 1.605               \\
                               & MIAS-LCEC        & 0.996                        & 0.182               & 0.659                           & 0.114               \\
                               & CLAIM            & 0.534                        & 0.065               & 0.376                           & \textbf{0.086}      \\ \hline
    \multirow{2}{*}{MF}        & EdO-LCEC         & 0.720                        & 0.106               & \textbf{0.349}                  & 0.109               \\
                               & CLAIM-4F         & \textbf{0.457}               & \textbf{0.046}      & 0.366                           & \textbf{0.086}      \\ \hline
    \end{tabular}
    \vspace{+0.2cm}
\end{table}

\subsection{Ablation Study}
To explore the contribution of each component in CLAIM, we further conduct an ablation study based on the setting in Sec. \ref{sec:exp_rough_initial}. The final results are presented in \tref{table:Exp4}, where CLAIM-S0 represents the given rough initial guess, CLAIM-S1, CLAIM-S2 correspond to the intermediate results after the grid search and the coarse random search. CLAIM-Structure and CLAIM-Texture means only the structure or texture loss is used in random searches. From the results, we can see how our pipeline obtains finer calibration results step by step, especially for the rotation (as shown in \fref{fig:waymo_4steps}). Meanwhile, it indicates that the structure loss solely can also help attain good calibration results while the texture loss is not robust enough when used separately.
After combining the two losses, more environmental information is incorporated and the accuracy can be further improved. We notice that the gain of texture loss for KITTI ($0.010^{\circ}$ and $0.008\mathrm{m}$) is subtle in contrast to that for Waymo ($0.084^{\circ}$ and $0.031\mathrm{m}$), which we consider is due to that KITTI's image height (375) is too small to capture abundant texture details. 
\begin{table}[]
    \centering
    \renewcommand{\arraystretch}{1.2}
    \caption{Ablation study of different components of CLAIM on the KITTI and Waymo dataset. The best results are show in bold.}
    \label{table:Exp4}

    \begin{tabular}{c|cc|cc}
    \hline
    \multirow{2}{*}{Methods} & \multicolumn{2}{c|}{KITTI}                         & \multicolumn{2}{c}{Waymo}                             \\
                             & $e_{r}\left(^{\circ}\right)$ & $e_{t}^+(\mathrm{m})$ & $e_{r}\left({ }^{\circ}\right)$ & $e_{t}^+(\mathrm{m})$ \\ \hline
    CLAIM-S0                 & 17.321                       & 0.346               & 17.321                          & 0.346               \\
    CLAIM-S1                 & 1.252                        & 0.346               & 1.750                           & 0.346               \\
    CLAIM-S2                 & 0.513                        & 0.115               & 0.323                           & 0.113               \\
    CLAIM                    & \textbf{0.472}               & \textbf{0.114}      & \textbf{0.253}                  & \textbf{0.099}      \\
    CLAIM-Structure          & 0.482                        & 0.122               & 0.337                           & 0.130               \\
    CLAIM-Texture            & 2.196                        & 0.391               & 0.710                           & 0.176               \\ \hline
    \end{tabular}
    \vspace{-0.2cm}
    \end{table}
    
\section{CONCLUSIONS}
We propose a novel target-free and feature-free method for LiDAR-camera alignment named CLAIM. Without complicated data processing and feature matching like previous methods, CLAIM uses a structure loss w.r.t. monodepth image and LiDAR depth projection in combination with a texture loss w.r.t. grayscale image and LiDAR intensity projection as the alignment metric. A coarse-to-fine searching method is utilized to find the optimal extrinsics. The experimental results on public datasets show that our method outperforms other state-of-the-art methods in both accuracy and robustness.   

\addtolength{\textheight}{-12cm}   





\bibliographystyle{IEEEtran}
\bibliography{IEEEexample}

\begin{thebibliography}{10}
\providecommand{\url}[1]{#1}
\csname url@rmstyle\endcsname
\providecommand{\newblock}{\relax}
\providecommand{\bibinfo}[2]{#2}
\providecommand\BIBentrySTDinterwordspacing{\spaceskip=0pt\relax}
\providecommand\BIBentryALTinterwordstretchfactor{4}
\providecommand\BIBentryALTinterwordspacing{\spaceskip=\fontdimen2\font plus
\BIBentryALTinterwordstretchfactor\fontdimen3\font minus
  \fontdimen4\font\relax}
\providecommand\BIBforeignlanguage[2]{{%
\expandafter\ifx\csname l@#1\endcsname\relax
\typeout{** WARNING: IEEEtran.bst: No hyphenation pattern has been}%
\typeout{** loaded for the language `#1'. Using the pattern for}%
\typeout{** the default language instead.}%
\else
\language=\csname l@#1\endcsname
\fi
#2}}

\bibitem{Hand-Eye}
R.~Horaud and F.~Dornaika, ``{Hand-Eye Calibration},'' \emph{The International
  Journal of Robotics Research}, vol.~14, pp. 195 -- 210, 1995.

\bibitem{Chessboard-points}
\BIBentryALTinterwordspacing
J.~Cui, J.~Niu, Z.~Ouyang, Y.~He, and D.~Liu, ``{ACSC: Automatic Calibration
  for Non-repetitive Scanning Solid-State LiDAR and Camera Systems},'' 2020.
  [Online]. Available: \url{https://arxiv.org/abs/2011.08516}
\BIBentrySTDinterwordspacing

\bibitem{Chessboard-Planes}
E.-s. Kim and S.-Y. Park, ``{Extrinsic Calibration between Camera and LiDAR
  Sensors by Matching Multiple 3D Planes},'' \emph{Sensors}, vol.~20, no.~1,
  2020.

\bibitem{ArUco-marker}
\BIBentryALTinterwordspacing
A.~Dhall, K.~Chelani, V.~Radhakrishnan, and K.~M. Krishna, ``{LiDAR-Camera
  Calibration using 3D-3D Point correspondences},'' 2017. [Online]. Available:
  \url{https://arxiv.org/abs/1705.09785}
\BIBentrySTDinterwordspacing

\bibitem{Wooden-board}
C.~Guindel, J.~Beltrán, D.~Martín, and F.~García, ``{Automatic extrinsic
  calibration for lidar-stereo vehicle sensor setups},'' in \emph{2017 IEEE
  20th International Conference on Intelligent Transportation Systems (ITSC)},
  2017, pp. 1--6.

\bibitem{HKU-Mars}
C.~Yuan, X.~Liu, X.~Hong, and F.~Zhang, ``{Pixel-level extrinsic self
  calibration of high resolution lidar and camera in targetless
  environments},'' \emph{IEEE Robotics and Automation Letters}, vol.~6, no.~4,
  pp. 7517--7524, 2021.

\bibitem{Line-zhang}
X.~Zhang, S.~Zhu, S.~Guo, J.~Li, and H.~Liu, ``{Line-based Automatic Extrinsic
  Calibration of LiDAR and Camera},'' in \emph{2021 IEEE International
  Conference on Robotics and Automation (ICRA)}, 2021, pp. 9347--9353.

\bibitem{CRLF}
\BIBentryALTinterwordspacing
T.~Ma, Z.~Liu, G.~Yan, and Y.~Li, ``{CRLF: Automatic Calibration and Refinement
  based on Line Feature for LiDAR and Camera in Road Scenes},'' 2021. [Online].
  Available: \url{https://arxiv.org/abs/2103.04558}
\BIBentrySTDinterwordspacing

\bibitem{Sun}
C.~Sun, Z.~Wei, W.~Huang, Q.~Liu, and B.~Wang, ``{Automatic Targetless
  Calibration for LiDAR and Camera Based on Instance Segmentation},''
  \emph{IEEE Robotics and Automation Letters}, vol.~8, no.~2, pp. 981--988,
  2023.

\bibitem{CalibAnything}
\BIBentryALTinterwordspacing
Z.~Luo, G.~Yan, and Y.~Li, ``{Calib-Anything: Zero-training LiDAR-Camera
  Extrinsic Calibration Method Using Segment Anything},'' 2023. [Online].
  Available: \url{https://arxiv.org/abs/2306.02656}
\BIBentrySTDinterwordspacing

\bibitem{CalibDepth}
J.~Zhu, J.~Xue, and P.~Zhang, ``{CalibDepth: Unifying Depth Map Representation
  for Iterative LiDAR-Camera Online Calibration},'' in \emph{2023 IEEE
  International Conference on Robotics and Automation (ICRA)}, 2023, pp.
  726--733.

\bibitem{CalibNet}
G.~Iyer, R.~K. Ram, J.~K. Murthy, and K.~M. Krishna, ``{CalibNet: Geometrically
  supervised extrinsic calibration using 3D spatial transformer networks},'' in
  \emph{2018 IEEE/RSJ International Conference on Intelligent Robots and
  Systems (IROS)}.\hskip 1em plus 0.5em minus 0.4em\relax IEEE, 2018, pp.
  1110--1117.

\bibitem{depthv2}
\BIBentryALTinterwordspacing
L.~Yang~et al., ``{Depth Anything V2},'' 2024. [Online]. Available:
  \url{https://arxiv.org/abs/2406.09414}
\BIBentrySTDinterwordspacing

\bibitem{Hand-Eye-method2}
N.~Ou, H.~Cai, J.~Yang, and J.~Wang, ``{Targetless Extrinsic Calibration of
  Camera and Low-Resolution 3-D LiDAR},'' \emph{IEEE Sensors Journal}, vol.~23,
  pp. 10\,889--10\,899, 2023.

\bibitem{MIAS-LCEC}
Z.~Huang, Y.~Zhang, Q.~Chen, and R.~Fan, ``{Online, target-free lidar-camera
  extrinsic calibration via cross-modal mask matching},'' \emph{IEEE
  Transactions on Intelligent Vehicles}, 2024.

\bibitem{EDO-LCEC}
\BIBentryALTinterwordspacing
Z.~Huang, J.~Li, P.~Zhong, and R.~Fan, ``{Environment-Driven Online
  LiDAR-Camera Extrinsic Calibration},'' 2025. [Online]. Available:
  \url{https://arxiv.org/abs/2502.00801}
\BIBentrySTDinterwordspacing

\bibitem{UMich}
G.~Pandey, J.~R. McBride, S.~Savarese, and R.~M. Eustice, ``{Automatic
  extrinsic calibration of vision and lidar by maximizing mutual
  information},'' \emph{Journal of Field Robotics}, vol.~32, no.~5, pp.
  696--722, 2015.

\bibitem{DVL}
K.~Koide, S.~Oishi, M.~Yokozuka, and A.~Banno, ``{General, Single-shot,
  Target-less, and Automatic LiDAR-Camera Extrinsic Calibration Toolbox},'' in
  \emph{2023 IEEE International Conference on Robotics and Automation (ICRA)},
  2023, pp. 11\,301--11\,307.

\bibitem{SparseGS}
\BIBentryALTinterwordspacing
H.~Xiong, S.~Muttukuru, R.~Upadhyay, P.~Chari, and A.~Kadambi, ``{SparseGS:
  Real-Time 360$^\circ$ Sparse View Synthesis using Gaussian Splatting},''
  2024. [Online]. Available: \url{https://arxiv.org/abs/2312.00206}
\BIBentrySTDinterwordspacing

\bibitem{3DGS}
B.~Kerbl, G.~Kopanas, T.~Leimk{\"u}hler, and G.~Drettakis, ``{3D Gaussian
  Splatting for Real-Time Radiance Field Rendering},'' \emph{ACM Transactions
  on Graphics}, vol.~42, no.~4, pp. 1--14, 2023.

\bibitem{KITTI}
A.~Geiger, P.~Lenz, and R.~Urtasun, ``{Are we ready for autonomous driving? The
  KITTI vision benchmark suite},'' in \emph{2012 IEEE Conference on Computer
  Vision and Pattern Recognition}, 2012, pp. 3354--3361.

\bibitem{Waymo}
P.~Sun~et al., ``{Scalability in Perception for Autonomous Driving: Waymo Open
  Dataset},'' in \emph{2020 IEEE/CVF Conference on Computer Vision and Pattern
  Recognition (CVPR)}, 2020, pp. 2443--2451.

\bibitem{Zhu}
Y.~Zhu, C.~Li, and Y.~Zhang, ``{Online Camera-LiDAR Calibration with Sensor
  Semantic Information},'' in \emph{2020 IEEE International Conference on
  Robotics and Automation (ICRA)}, 2020, pp. 4970--4976.

\end{thebibliography}

\end{document}